\documentclass[letterpaper, 10 pt, conference]{ieeeconf}  

\IEEEoverridecommandlockouts                              

\newcommand{\RNum}[1]{\uppercase\expandafter{\romannumeral #1\relax}}

\usepackage{graphics} 
\usepackage{graphicx}
\graphicspath{{./figure/}}
\usepackage{subcaption}
\usepackage[font=small]{caption}

\usepackage{bm}

\usepackage[noadjust]{cite}

\usepackage{hyperref}
\usepackage{amsmath} 

\DeclareMathOperator*{\argmin}{argmin} 
\DeclareMathOperator*{\argmax}{argmax}

\usepackage{amssymb}  
\usepackage{algorithm, algpseudocode}

\usepackage{booktabs}
\usepackage{multirow}
\usepackage{cleveref}
\usepackage{color}

\usepackage{array}
\newcolumntype{C}[1]{>{\centering\arraybackslash}m{#1}}
\newcolumntype{N}{@{}m{0pt}@{}}

\usepackage{tabu}
\usepackage{siunitx}

\title{\LARGE \bf
	Effective Estimation of Contact Force and Torque for Vision-based Tactile Sensor with Helmholtz-Hodge Decomposition
}

\author{Yazhan Zhang$^{1}$, Zicheng Kan$^{1}$, Yang Yang$^{1}$, \textit{Member, IEEE}, Yu Alexander Tse$^{1}$ \\
	and Michael Yu Wang$^{2}$, \textit{Fellow, IEEE}
	\thanks{*Research is supported by the Hong Kong Innovation and Technology
Fund (ITF) ITS-018-17FP.}
	\thanks{$^{1}$Y. Zhang, Z. Kan, Y. Tse and Y. Yang are with the Department of
Mechanical and Aerospace Engineering, Hong Kong University of Science
and Technology, Hong Kong (e-mail: yzhangfr@connect.ust.hk; zkan@connect.ust.hk; yatse@connect.ust.hk; rayang@ust.hk).}%
	\thanks{$^{2}$M. Y. Wang (corresponding author) is with the Department of Mechanical
and Aerospace Engineering and the Department of Electronic and Computer
Engineering, Hong Kong University of Science and Technology, Hong Kong
(tel.: +852-34692544; e-mail: mywang@ust.hk).}%
}

\begin{document}

\maketitle
\thispagestyle{empty}
\pagestyle{empty}

\begin{abstract}

Retrieving rich contact information from robotic tactile sensing has been a challenging, yet significant task for the effective perception of object properties that the robot interacts with. This work is dedicated to developing an algorithm to estimate contact force and torque for vision-based tactile sensors. We first introduce the observation of the contact deformation patterns of hyperelastic materials under ideal single-axial loads in simulation. Then based on the observation, we propose a method of estimating surface forces and torque from the contact deformation vector field with the Helmholtz-Hodge Decomposition (HHD) algorithm. Extensive experiments of calibration and baseline comparison are followed to verify the effectiveness of the proposed method in terms of prediction error and variance. The proposed algorithm is further integrated into a contact force visualization module as well as a closed-loop adaptive grasp force control framework and is shown to be useful in both visualization of contact stability and minimum force grasping task.


\end{abstract}


\section{INTRODUCTION}

Tactile sensing has been investigated and proven to play critical roles in human interaction with the environment. For a robotic system, tactile sensor is also a key component for its perception system, especially in contact-rich manipulation tasks. However, tactile sensing technologies are relatively unexplored, comparing with great attention drawn to studies of visual perception principles and developments of algorithms in these decades, despite its complementary role to visual sensing in robotic scene perception.

The past few decades have seen vast emergence of various types of tactile sensors with different transducing principles, including capacitive, piezoelectric, piezoresistive, magneto-electric, etc. \cite{Dahiya:2010:TSH:1771964.1771965}. Recently, vision-based tactile sensors have been thriving and appearing in various robotic systems with advantages of easy fabrication, high resolution, and multi-axial deformation sensing capability, e.g. Gelforce \cite{sato2010finger}, FingerVision \cite{yamaguchi2016combining}, Gelsight \cite{yuan2015measurement} and a more compact Gelslim in \cite{donlon2018gelslim}. 
In our previous work, we have developed a vision-based tactile sensor also called FingerVision \cite{zhang2018fingervision} (the name FingerVision was first introduced in \cite{yamaguchi2016combining}) and it was proven to be effective in the slip detection task. In this work, we aim at further exploiting the capability of recovering contact force and torque from the displacement field of the vision-based tactile sensor.

\begin{figure}
	\centering
	\includegraphics[width=0.4\textwidth]{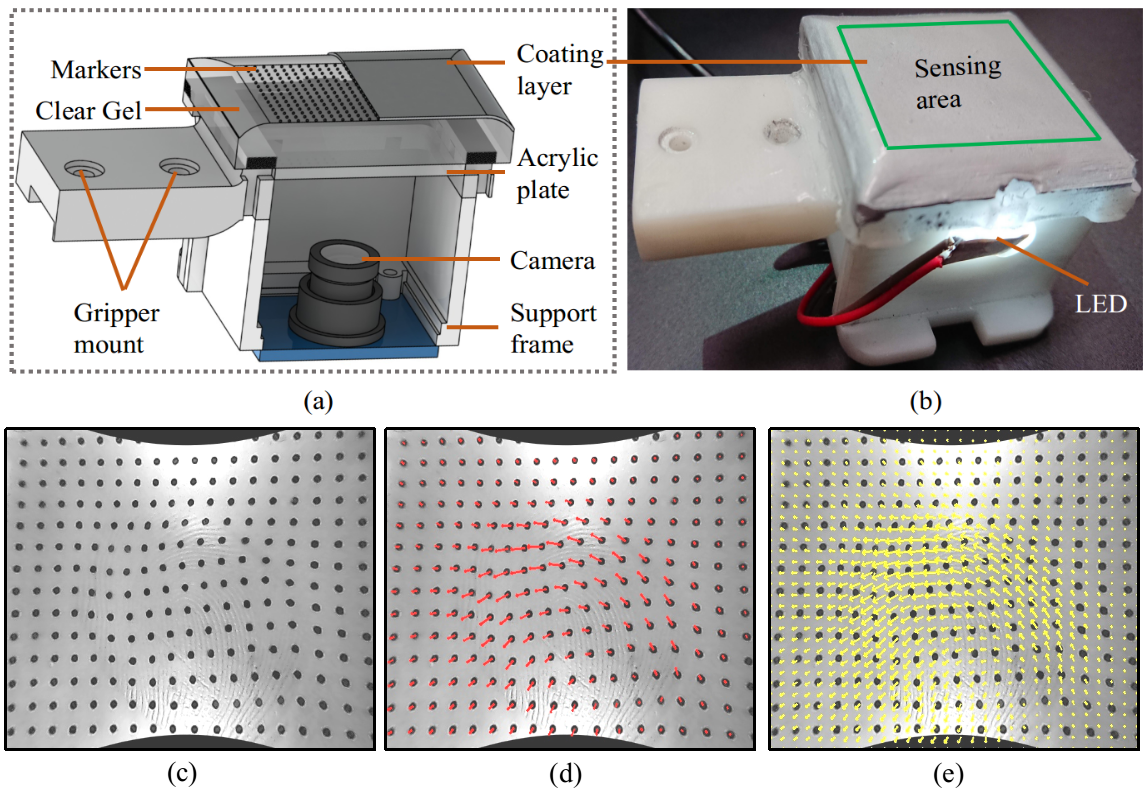}
	\caption{FingerVision tactile sensor. (a). Rendered 3D model  (in cutaway view). (b). Sensor prototype (details are referred to \cite{zhang2018fingervision}). (c-e). Raw image obtained from sensor, image with tracked displacement vectors and image with grid interpolated displacement vectors overlaying on top, respectively.}
	\label{fingervision_model}
\end{figure}

There are various ways to encode tactile signals, among which contact force and torque estimation from raw tactile information is of special interest, for it directly relates to statics or dynamics of the object during the interaction. For instance, human's intuitive feeling of finger skin traction and pressure and estimation of the center of mass of objects greatly enhance the success rate of dexterous, dynamic manipulation. In a robotic system, similarly, accurate force feedback helps robot capture the motion of the object and state transitions including contact making, slipping and contact breaking. Therefore, it endows robots with the capability of assessing grasp stability, which is essential for the successful execution of complex manipulation tasks.

For FingerVision sensor we developed in \cite{zhang2018fingervision} as reprinted in Fig. \ref{fingervision_model}, the sensing body is a clear elastomer with embedded black markers used as vision tracking features. The marker displacement vectors are seen as the grid sampling of the deformation in the elastomer layer.
When applied with external force and torque, deformation occurs in the hyperelastic body of FingerVision following continuum mechanics and the deformation fields show corresponding patterns under specific single-axial surface force and torque. 
Bearing this hint, decomposition of raw displacement vector field into multiple separated vector fields with specific patterns would be potentially helpful to decouple the deformation under multi-axial loads. However, the displacement field patterns are also correlated with shapes of contact area and affected by nonlinear deformation induced by large contact forces and torques, and interference of force and torque between axes. Thus, evaluation of method's generalization capability and proper selection of working range are necessary.

In this paper, our goal is to effectively recover the contact surface force and torque from vision-based tactile sensors. Toward this target, these contributions are generated in our work:

\begin{itemize}
	\item Introduction of the displacement field patterns of elastomer on the vision-based tactile sensor when applied with single-axial forces and their quantitative properties are presented.
	\item Proposal of a method to decompose displacement field of vision-based sensors into components that can be further used in estimating contact force and torque based on Helmholtz-Hodge Decomposition algorithm. The proposed method is both data efficient and with low model complexity for regression.
\end{itemize}

The rest of this paper is arranged in the following structure: Section \RNum{2} introduces previous works related to methods of force estimation for tactile sensors. In section \RNum{3}, we explain the patterns observed and formulate mapping functions from vector fields of specific patterns to corresponding contact forces in simulation. Afterwards, we propose that HHD algorithm can be used to decompose displacement vector field into components with similar patterns that leads to estimation of contact force and torque. In section \RNum{4}, extensive characteristic experiments and comparison to state-of-the-art methods are given to show the effectiveness of the proposed method. In section \RNum{5}, we integrate the proposed method into a contact stability visualization and grasping force feedback control framework. Finally, discussion and conclusion are drawn in section \RNum{6}.


\section{Related Works}


\subsection{Tactile Sensors and Force Measurement}

Vision-based tactile sensors attract increasing attention for its sensing capability with multi-modal contact information in addition to advantages of superior sensing resolution, including deformation \cite{sato2010finger}\cite{yamaguchi2016combining}, object texture \cite{johnson2009retrographic,dong2017improved,yuan2017connecting}, contact area estimation \cite{donlon2018gelslim}, geometry reconstruction \cite{johnson2009retrographic}\cite{yuan2017gelsight} and force estimation \cite{sato2010finger}\cite{yamaguchi2016combining}\cite{yuan2017gelsight}\cite{mcinroe2018towards}. Besides, vision-based tactile sensors have been shown to perform well in high-level tasks like object recognition \cite{yuan2017connecting}, localization of dynamic object \cite{li2014localization}, and slip detection \cite{yuan2015measurement}\cite{zhang2018fingervision}\cite{dong2017improved}\cite{van2018slip}. Surface deformation serves as a basic signal modality for above higher-level information in these sensing systems.

\begin{figure}
	\centering
    \includegraphics[width=0.35\textwidth]{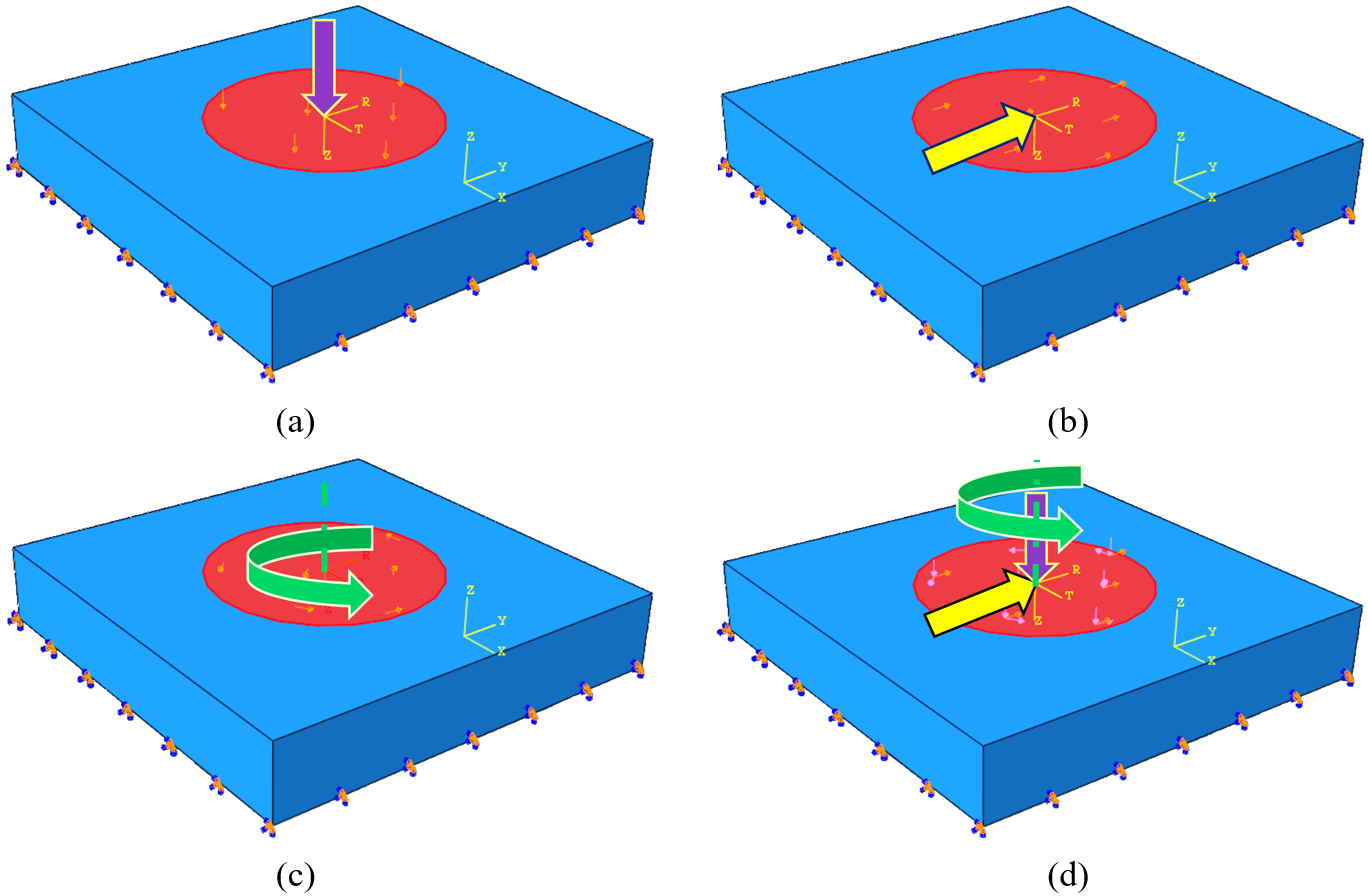}
	\caption{Applied contact force configurations in simulation. (a). Normal force distributed uniformly. (b). Unidirectional tangential force distributed uniformly. (c). Torsional force along normal axis. (d). Combination of tangential, normal, and torsional forces.}
\label{force_configuration}
\end{figure}

Since the contact deformation is only one of the intermediate states for robotic manipulation feedback loop, researchers have been putting efforts into developing methods for recovering contact forces for tactile sensors. Generally, contact pressure distribution is relatively easier to be extracted for traditional capacitive, piezoelectric tactile array \cite{Dahiya:2010:TSH:1771964.1771965} or sensors utilizing total internal reflective (TIR) principle as presented in \cite{begej1988planar}. However, multi-axial-force estimation is much more challenging by comparison.
 Ohka et al. \cite{ohka1996data} presented a tactile sensor made of a rubber layer and a pyramid-shaped indenter on an acrylic plate that was able to capture changes of indentations of the pyramid array into the rubber skin with camera. According to the changes of the indentation areas, they successfully predicted three-axial contact forces.  
 Sato et al. \cite{sato2010finger} fabricated a vision-based sensor called Gelforce with double-layer markers in different colors as tracking targets, which enables the measurement of motion along the surface normal via tracking the movement differences between markers in two layers. Based on an observational method and calibration, multi-axis force could be extracted from this complex fingertip-shaped sensor. Calibration procedures were specifically designed for the sensors making contact with probe-shape objects and generalization testing to different contact objects were not performed.
  Vogt et al. \cite{vogt2013design} built a microfluid-based flexible skin that can detect and differentiate normal and shear force, whereas the system suffered from a lower response time that was not suitable for robotic scenarios. In addition, the microfluid-based sensor could only estimate force and was inferior in multi-modality sensing by comparison with vision-based tactile sensors.
  
  Neural network has shown its usefulness in recovering contact force for tactile sensors.
  Maria et al. \cite{de2012force} designed a tactile sensor using an array of paired light emitters and receivers that was able to capture deformation in local region and infer contact forces with trained neural network. 
  In \cite{fang2018dual}, multi-layer neural network was utilized in mapping from markers displacement field to three-axial contact force with a relatively low error on a Gelsight-like sensor. However, neural networks are usually not data-efficient and suffers from overfitting when only a small amount of data is available. Additionally, above works also didn't discuss the generalization performance on different contact objects. 
  In our work, we start by observing the response patterns of displacement field to different force and torque configurations, and based on the observation, we decompose vector field into components containing individual patterns to infer decoupled contact forces. This method significantly reduces the dimension of deformation vector field and is shown to retain good invariance to different contact objects. 

\subsection{Helmholtz-Hodge Decomposition}

Helmholtz-Hodge Decomposition is commonly used in motion analysis, e.g. target tracking in computer vision, computational fluid motions analysis \cite{guo2005efficient}, acting as feature extraction to capture divergence source, sink and vertex of rotational motion for vector fields. HHD describes a vector field in the form of the summation of a divergence-free, a curl-free, and a harmonic flow, with manually set boundary condition imposed to get a unique solution. In \cite{bhatia2014natural}, Bhatia et al. proposed a natural Helmholtz-Hodge Decomposition (nHHD) method enabling defect-free analysis for various boundaries conditions with a data-driven method. In this work, we adopt nHHD to decompose our displacement field into separated components corresponding to the responses of specific external contact forces. We show that this tool is effective in recovering contact forces for the deformable medium used in most vision-based tactile sensors by quantitative analysis, although theoretical relations between the decomposition component patterns and patterns observed in the simulation have not been established yet.


\section{Method Description}

\begin{figure}
	\centering
	\includegraphics[width=0.48\textwidth]{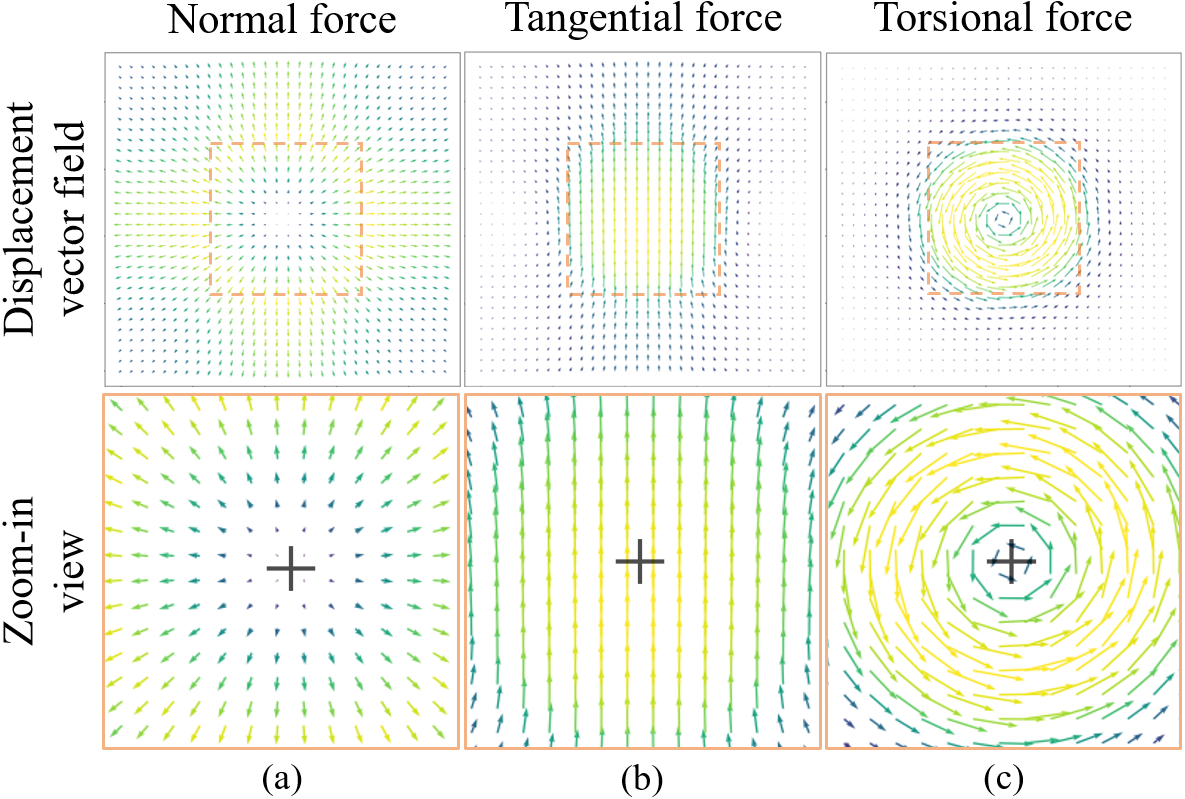}
	\caption{Displacement fields of elastomer body under three load configurations shown in Fig. \ref{force_configuration}(a-c).}
	\label{response_simulation}
\end{figure}

Vision-based tactile sensors, such as Gelforce, FingerVision, Gelsight, make use of the deformation captured by the camera to infer contact forces by following hyperelastic continuum mechanics \cite{sato2010finger}, data fitting with calibration \cite{de2012force}\cite{fang2018dual} or both combined \cite{sato2010finger}. For analysis of hyperelastic deformation, finite element method (FEM) is commonly used. FEM approximates stress and strain response under external force that governed by continuum mechanics with finite number of nodes. To obtain an accurate result of surface motion, it is a common practice to increase the number of nodes with a proper meshing method, which results in increased dimension of the stiffness matrix that might be over demanding for computation in real-time applications. In this work, we take advantage of the insight that the displacement fields of the elastomer show unique and consistent graphical patterns under different single-axial loads (normal, tangential, and torsional loads) in simulation. These patterns possess quantitative properties that can be leveraged to formulate mapping from vector field with patterns to contact forces.

\subsection{Behavior under Different Loads}

For contact in reality, any surface traction comes in the form of contact friction, and thus tangential force would not exist without normal pressure being applied simultaneously. To explore the behavior of the displacement field under loads along different axes separately, we simulate with hyperelastic material in Abaqus. As shown in Fig. \ref{force_configuration}, within circle region on the top (in red), uniformly distributed normal force, tangential force and torsion along the surface normal (directions are shown with arrows) are applied with fixed bottom faces as boundary conditions. With these three configurations, typical simulation results are shown in Fig. \ref{response_simulation}. The displacement vector fields are obtained by further interpolating on a fixed-spacing grid and rendered with colors coding vectors' magnitude.

\begin{figure}
	\centering
	\includegraphics[width=0.44\textwidth]{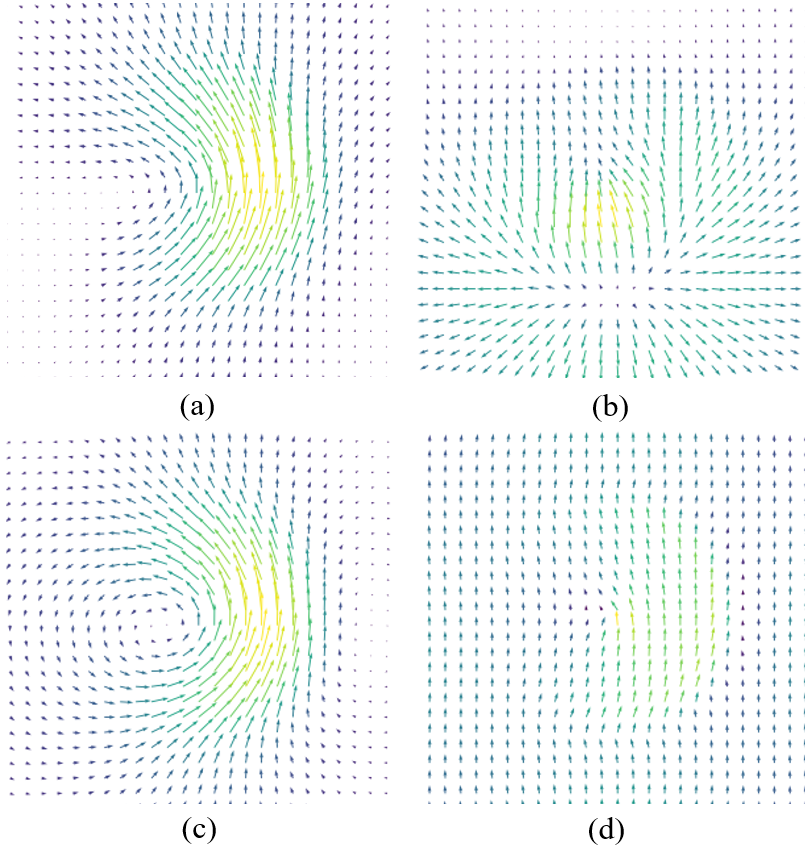}
	\caption{Decomposition result of simulated displacement field with the multi-axial loads. (a). Displacement vector field. (b). Curl-free component. (c). Divergence-free component. (d). Harmonic component.}
	\label{decomposition_simulation}
\end{figure}

Let $\vec{v}_i = \left( \delta x_i, \delta y_i \right) $ denotes $i^{th}$ displacement vector and $ V_i=\left\lbrace p_i, \vec{v}_i \right\rbrace $ denote $i^{th}$ displacement vector associated with position $ p_i = \left( x_i, y_i\right)$ being the start of the vector. Assume that the rotational centers $c$ of configuration are known, let $\vec{r}_{ij}$ be the arm of moment of $i^{th}$ vector w.r.t $j^{th}$ rotation center. Let $\vec{r}^a_i$ be the arm of moment w.r.t divergence center (the cross in Fig. \ref{response_simulation}(a)) and $\vec{r}^b_i$ be the arm of moment w.r.t the contact center (location of vector with maximum magnitude, the cross in Fig. \ref{response_simulation}(b)). From the displacement vector fields in simulation, it is observed that three graphical patterns of divergence, unidirection and rotation can be generated under normal, tangential and torsional forces correspondingly. With these patterns in Fig. \ref{response_simulation}, we notice the following quantitative properties:

\begin{itemize}
    \item For pattern (a), norm of vector summation $\smash{|\sum^{N-1}_{i=0} \vec{v}_{i}|}$ and magnitude of summation of moments w.r.t. the divergence center $\smash{\sum ^{N-1}_{i=0} \vec{r}^a_{i} \times \vec{v}_{i}}$ both yield small values, while summation of vector norms $\smash{\sum^{N-1}_{i=0} |\vec{v}_{i}|}$ gives a significantly larger value.
    
    \item For pattern (b), summation of moments w.r.t. the contact center $\smash{\sum ^{N-1}_{i=0} \vec{r}^b_{i} \times \vec{v}_{i}}$ yields a small magnitude, while norm of vector summation  $\smash{|\sum^{N-1}_{i=0} \vec{v}_{i}|}$ gives a larger value by comparison.
    
    \item For pattern (c), norm of vector summation $\smash{|\sum^{N-1}_{i=0} \vec{v}_{i}|}$ yields a small value, while summation of moments w.r.t the rotational center $\smash{\sum^{M}_{j} \sum ^{N-1}_{i=0} \vec{r}_{ij} \times \vec{v}_{i}}$ gives a much larger magnitude.
\end{itemize}
where N is the number of vectors, and M is the number of rotational centers of the vector field.

Assuming that an arbitrary vector field $\vec{V}$ is composed of vector fields with these diverging $\vec{V}_n$, unidirectional $\vec{V}_t$, and rotational $\vec{V}_{\tau}$ patterns, and following the quantitative properties above, we have formulations below

\begin{equation}
\begin{aligned}
    S_n &= \sum^{N-1}_{i=0, \vec{v}_i \in \vec{V}_n} |\vec{v}_{i}| = \sum^{N-1}_{i=0, \vec{v}_i \in \vec{V}_n + \vec{V}_\tau} |\vec{v}_{i}|\\
    S_t &= |\sum^{N-1}_{i=0, v_i \in \vec{V}_t} \vec{v}_{i}| = |\sum^{N-1}_{i=0, \vec{v}_i \in \vec{V}} \vec{v}_{i}|\\
    S_\tau &= \sum^{M}_{j} \sum ^{N-1}_{i=0, \vec{v}_i \in \vec{V}_\tau} \vec{r}_{ij} \times \vec{v}_{i}= \sum^{M}_{j} \sum ^{N-1}_{i=0, \vec{v}_i \in \vec{V}_n + \vec{V}_\tau} \vec{r}_{ij} \times \vec{v}_{i}\\
\end{aligned}
\label{eq: three_values}
\end{equation}
where $\vec{V} = \vec{V}_n + \vec{V}_t + \vec{V}_\tau$, and $S_n$, $S_t$ and $S_\tau$ are summation of vector norms on $\vec{V}_n$, norm of vector summation on $\vec{V}$, and total moments of vectors w.r.t. rotational centers on $\vec{V}_\tau$.

In reverse, estimation of contact force and torque can follow the scheme of computing $S_t$ given a displacement field $\vec{V}$ first, then decomposing $\vec{V}$ into $\vec{V}_n$ and $\vec{V}_\tau$ for computation of $S_n$ and $S_\tau$ following Eq. (\ref{eq: three_values}). The problem boils down to finding a suitable decomposition method.

\subsection{Decomposition Algorithm}

HHD method is a tool widely used in flow physics analysis to gain insights into such features as critical points, divergence source, sink, rotational vertex and curl distribution, etc.\cite{guo2005efficient}. H. Bhatia et al. \cite{bhatia2014natural} presented a natural HHD (nHHD) to tackle data-dependent boundary condition selection problem. In our work, we adopt nHHD to compute separated vector fields for the reason that in a 2D space, the displacement of elastomer under torsional and normal loads from simulation results have similar pattern representations to that in divergence-free and curl-free fields decomposed by nHHD. 

According to \cite{bhatia2014natural}, considering the above smooth displacement vector field $\vec{V}: \mathbb{\si{\ohm}} \rightarrow \mathbb{R}^n$, where $\mathbb{\si{\ohm}} \subseteq \mathbb{R}^n$ (e.g. n = 2 in 2D case), we have

\begin{equation}
	\vec{V} =\vec{d} \ +\ \vec{r} \ +\vec{h}
	\label{hdd_decomposition}
\end{equation}
where $\vec{d}$ denotes curl-free component ($\nabla \times \vec{d} =\vec{0}$), $\vec{r}$ is divergence-free component ($\nabla \cdot \vec{r} =0$) and $\vec{h}$ is harmonic ($\nabla \times \vec{h} =\vec{0}, \nabla \cdot \vec{h} =0$). Eq. (\ref{hdd_decomposition}) is further transformed into Eq. (\ref{hdd_potential}) in the form of gradients of two scalar potential functions $R$ and $D$

\begin{equation}
	 \begin{split}
	\vec{V} & =\nabla D+\nabla \times R +\vec{h}\\
	& =\nabla D+J\nabla R+\vec{h}
	\end{split}
	\label{hdd_potential}
\end{equation}
where $\vec{d} \ =\ \nabla D$ and $\vec{r} =\nabla \times R =J\nabla R$ with $J$ being the $\pi/2$-rotation matrix. By applying divergence and curl operations, we obtain following Poisson equations

\begin{equation}
 \begin{array}{l}
\Delta D=\nabla \cdot \vec{V}\\
\Delta R=-\nabla \cdotp J\vec{V} 
\end{array}
\label{poisson_function}
\end{equation}

 Therefore, Eq. (\ref{poisson_function}) can be solved using Green's function in the domain to obtain $\vec{d}$ and $\vec{r}$, and data-dependent boundary conditions are imposed to derived harmonic component $\vec{h}$ uniquely. For more implementation details of the solving process, it is recommended to refer to \cite{bhatia2014natural}. The rotational centers $c_+$ and(or) $c_-$ in $\vec{r}$ are localized where the maxima and(or) the minima of $R$ are achieved over the discrete domain of $\vec{V}$ if the extrema exist, and the arms of moments $\vec{r}_{i+}$ and(or) $\vec{r}_{i-}$ of $i^{th}$ vector can be obtained as Eq. (\ref{eq:arm_of_moment}) presents.

\begin{equation}
\label{eq:arm_of_moment}
\begin{aligned}
    c_+ &= p_{i_{max}} | i_{max} = \argmax_i R_i \\
    c_- &= p_{i_{min}} | i_{min} = \argmin_i R_i\\
    \vec{r}_{i+} &= p_i - c_+\\
    \vec{r}_{i-} &= p_i - c_- 
\end{aligned}
\end{equation}
where $R_i$ is the value of the potential function $R$ at $p_i$.

\begin{figure}
	\centering
	\includegraphics[width=0.3\textwidth, height=3.5cm]{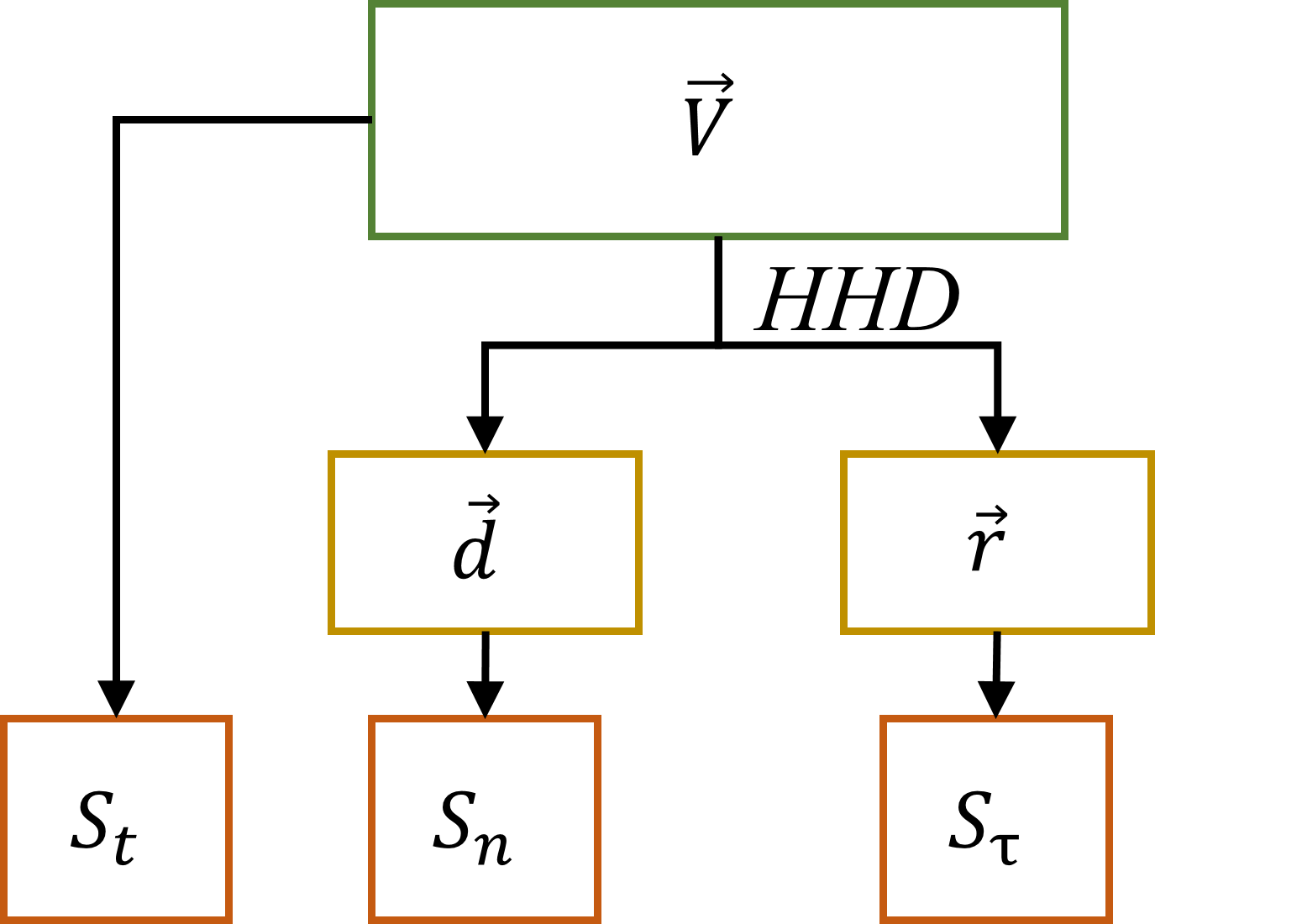}
	\caption{Contact force and torque computation pipeline.}
	\label{calculation_flow}
\end{figure}

The result of HHD for the simulated displacement field under multi-axial loads is given in Fig. \ref{decomposition_simulation}. Although it is noticeable that the patterns given in the separated components are not identical to those shown in Fig. \ref{response_simulation} in terms of distribution of vector magnitude, calculation of $S_n$, $S_t$, and $S_\tau$ remains valid according to Eq. (\ref{eq: three_values}).
By combining the procedure proposed in section \RNum{3}-A and nHHD algorithm, we present the procedure to compute $S_n$, $S_t$ and $S_\tau$. $S_t$ is obtained from the raw displacement field following Eq. (\ref{eq: three_values}), and in parallel, the raw displacement field is fed into HHD module to generate two fields of interests: curl-free and divergence-free fields. $S_n$ and $S_\tau$ are calculated from these two vector fields with Eq. (\ref{eq: three_values}) and Eq. (\ref{eq:arm_of_moment}). The calculation scheme is illustrated in Fig. \ref{calculation_flow}.

Let the mapping from $S_n$, $S_t$ and $S_\tau$ to contact normal force $F_n$, tangential force $F_t$ and torque along surface normal $F_\tau$ be functions $g_n$, $g_t$ and $g_\tau$, respectively which connect to the estimations of contact force and torque in Eq. (\ref{eq: mapping_functions}). With the significantly dimensional reduction from tactile displacement vector field to $S_n$, $S_t$ and $S_\tau$, it can be expected that the complexity of the model used to predict contact force and torque using decomposed results will be much lower, compared to that  using raw displacement field.  

\begin{equation}
    \begin{aligned}
        F_n &= g_n(S_n)\\
        F_t &= g_t(S_t)\\
        F_\tau &= g_\tau(S_\tau)
    \end{aligned}
    \label{eq: mapping_functions}
\end{equation}


\section{Experiments and Evaluation}

This section gives description of the characteristic experiments for the proposed decomposition algorithm including mapping function calibration and baseline comparison to evaluate the advantages of the decomposition method, compared with the method taking raw tactile displacement field as input. 

\subsection{Mapping Function Calibration}

\begin{figure}
	\centering
	\includegraphics[width=0.44\textwidth]{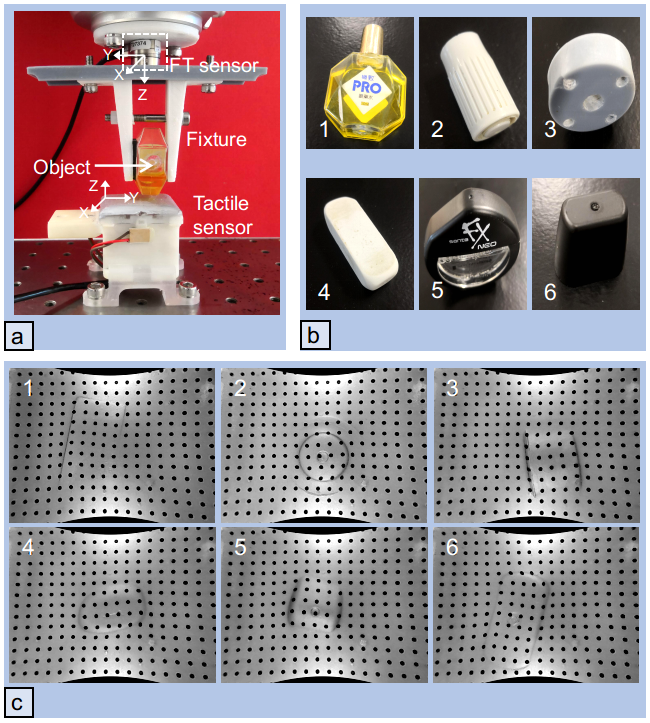}
	\caption{Experimental setup of data collection for calibration and baseline comparison. (a). Tactile image and contact force/torque collection with robot-arm-driven fixture fixing object to make contact on the tactile sensor. (b). 6 objects for contact making. (c). Examples of contact images for 6 objects.}
	\label{calibration_setup}
\end{figure}

Calibration is performed to find the mapping functions $g_n$, $g_t$ and $g_\tau$. Here we choose regression using a small amount of data, considering the dimensional reduction that our algorithm realizes. We collect force and torque data using highly accurate Force/Torque sensor depicted in Fig. \ref{calibration_setup}a. ATI nano-17 is installed on and driven by the UR10 robot arm. A simple 3D-printed gripper is mounted on tool side of the Force/Torque sensor and used as a fixture of objects. To examine the consistency between and generalization capability to a wide range of objects with different shapes, sizes, textures, hardness and elasticity, 6 objects with these variances are selected, as given in Fig. \ref{calibration_setup}b. when collecting dataset, objects are tightly grasped by the gripper and pressed onto the sensing surface of the tactile sensor. Typical tactile deformation images are shown in Fig. \ref{calibration_setup}c, with the corresponding object labels. 

\begin{figure*}
	\centering
	\includegraphics[width=0.95\textwidth]{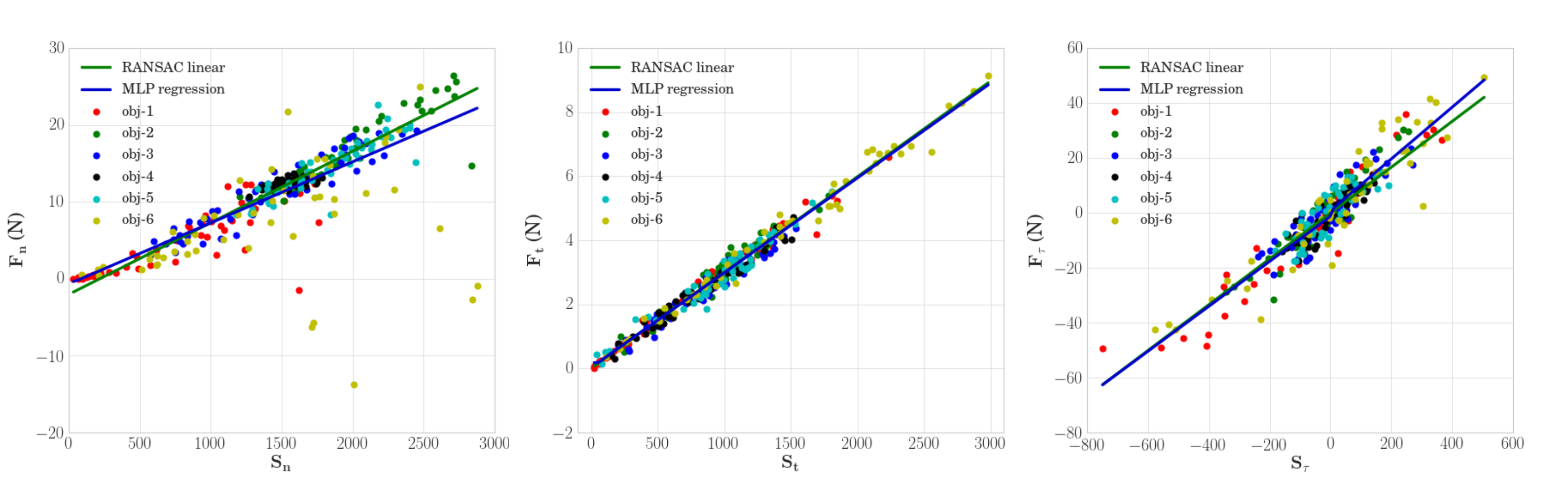}
	\caption{Calibration data and fitting results. Data collected using different objects are scattered with different colors. Data regression methods include RANSAC linear model and MLP regression. From left to right, charts are $F_n$ vs. $S_n$, $F_t$ vs. $S_t$ and $F_\tau$ vs. $S_\tau$.}
	\label{fig: char_plots}
\end{figure*}

As for the size of the calibration dataset, a total amount of 300 data points are collected, with 50 for each object. For every object, motions along  surface normal and tangent include pressing, surface dragging and twisting with randomized distance and angles in every data collection trial. The ranges of these randomized distance and angles are carefully adjusted to fit the working range of the sensor without dealing damage or too much wearing to the elastomer. 

The calibration data is presented in Fig. \ref{fig: char_plots},  calculations of $S_n$, $S_t$ and $S_\tau$ use the calculation pipeline in Fig. \ref{calculation_flow}. Qualitatively, the linearity of the data is strong in the selected working range, which achieves dimensional reduction and guarantees low complexity for models to approximate the distribution of data. Besides, we notice that for normal force, the data distribution is less concentrated compared with those of the other two sets of data. It is ascribed to the lack of capability of the monocular camera inside the sensor to capture the markers' motion along the sensor surface normal, in which direction the deformation of elastomer balances a  large portion of external normal force. As a result, only divergence motion in 2D plane is used for calculating normal force, leading to a larger variance in the distribution for the normal force data subset.

Two models with low complexity is fitted to the three sets of data. First, linear model with RANSAC outlier rejection algorithm is chosen, considering there exist some outliers in the data. For example, some of the measured normal forces are of negative values, which is impossible in common cases. RANSAC iteratively chooses group of inliers that lead to the lowest regression error. Second model is a three-layer multi-layer perceptron (MLP) that was used in the previous works \cite{de2012force}\cite{fang2018dual}. Since the underlying distribution is of relative low dimension, the regression of MLP to the data can also generate good performance when small model is applied given small amount of data. As shown in Fig. \ref{fig: char_plots}, RANSAC linear model and MLP model have close prediction values, except that RANSAC performs better in capturing underlying distribution in normal force case by rejecting outliers from object 1 and object 6.

\subsection{Baseline Comparison}

Baseline comparison is given in this section. Regarding the prediction performance of contact force and torque based on $S_n$, $S_t$ and $S_\tau$ calculated from the decomposed components and raw displacement vector fields, three regression models are compared with mean-square-error (RMSE) metric. For decomposed 1D data fitting, RANSAC linear model, and MLP regression with three-layer structure with 10 hidden units are adopted. As for input vectors without decomposition, a significantly larger MLP regressor with five-layer structure and $512 \times 128 \times 10$ hidden units is used.

Two MLP models are all fully trained with L-BFGS optimizer. 6-fold cross validation with splits of the data from different objects are used for evaluation of the overall performances of models and also biases toward certain objects. The results are shown in  Table \ref{tab:RMSE}. RANSAC linear model excels in terms of the mean  RSME of the prediction for normal force and tangential force cases, whereas MLP regressor on the 1D data performs better in the aspect of the prediction variances and slightly better in the mean RSME for torsional case. This could be attributed to the outlier rejection mechanism of RANSAC linear model to sustain the disturbance of noises, which lead to lower average prediction errors. As expected, all three models give a larger RSME in the normal force case when being evaluated on data collected with object 6 after being trained on the other 5 objects during cross validation. However, the combination of raw displacement vector with complex MLP gives a lower variance in this case, showing more consistent performance across different objects and with noises. In summary, linear models with decomposition capture the underlying distribution better given small amount of available data, while one can expect MLP without decomposition can improve highly if large dataset is collected.

\begin{table}[]
\centering
\caption{RMSE of different methods on estimating the contact force and torque based on the decomposed deformation vector fields or raw deformation one.}
\label{tab:RMSE}
\resizebox{0.45\textwidth}{!}
{%
\begin{tabular}{c|c|c|c|c}
\hline
\multicolumn{2}{c|}{RMSE} & \multicolumn{2}{c|}{Decomposition} & No decomposition \\ \hline
\multicolumn{2}{c|}{Method} & \begin{tabular}[c]{@{}c@{}}RANSAC \\ Linear\end{tabular} & \begin{tabular}[c]{@{}c@{}}MLP \\ Regression\end{tabular} & \begin{tabular}[c]{@{}c@{}}MLP \\ Regression\end{tabular} \\ \hline
\multicolumn{2}{c|}{\begin{tabular}[c]{@{}c@{}}Model \\ complexity\end{tabular}} & 2 & 10 & $512\times128\times10$ \\ \hline
\multirow{2}{*}{Normal (N)} & Mean & \textbf{2.952} & 3.286 & 4.482 \\
 & Stdv & 2.584 & 2.497 & \textbf{1.295} \\ \hline
\multirow{2}{*}{Tangential (N)} & Mean & \textbf{0.241} & 0.242 & 1.544 \\
 & Stdv & 0.033 & \textbf{0.032} & 0.813 \\ \hline
\multirow{2}{*}{Torsional (Nmm)} & Mean & 5.862 & \textbf{5.621} & 6.769 \\
 & Stdv & 1.547 & \textbf{1.353} & 3.621 \\ \hline
\end{tabular}%
}
\end{table}


\section{Grasping Tasks}

\begin{figure}
	\centering
	\includegraphics[width=0.485\textwidth]{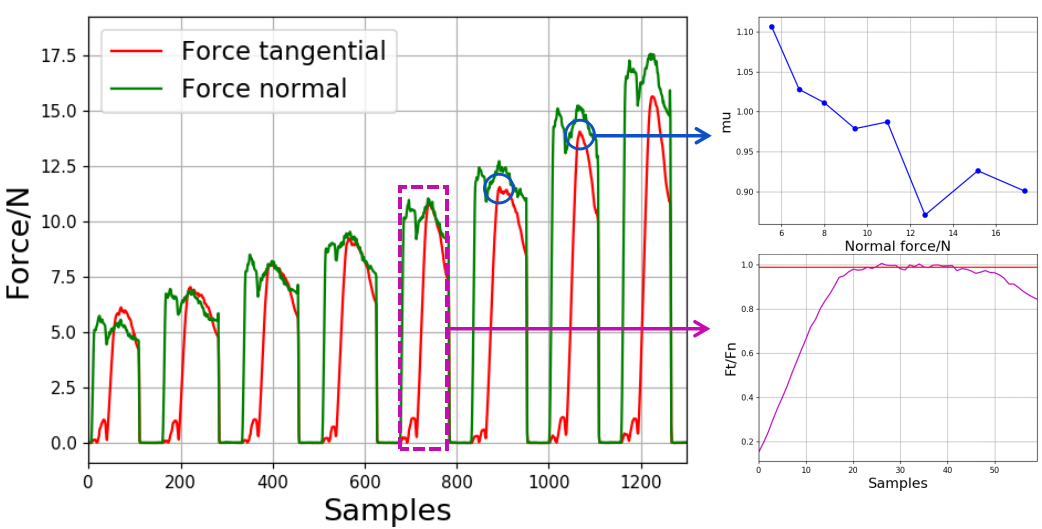}
	\caption{Contact force signals under multiple sliding motion trials. Upper right chart is generated by measuring ratios $F_t/F_n$ at peaks of $F_t$ (as blue circles marked). Lower right chart is the ratio $F_t/F_n$ inside the window delineated in purple dash rectangle.}
	\label{friction_data}
\end{figure}

\begin{figure}
	\centering
	\includegraphics[width=0.35\textwidth]{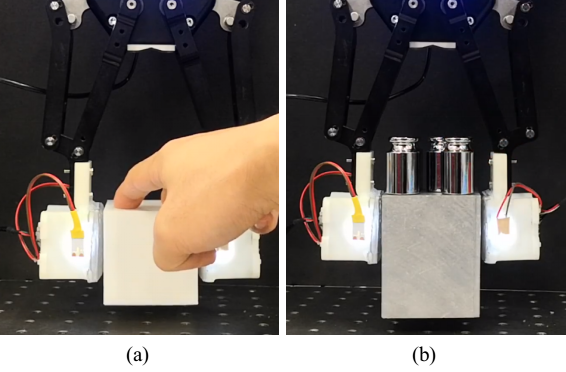}
	\caption{Adaptive grasping force control experiment. (a). Robotiq 2-finger 140 gripper with FingerVision as fingertips holding an object and then the object is pressed by hand till slip occurs. (b). Gripper with FingerVision holds an object and then the tangential load is increased/decreased by loading and unloading weights on top of the object. }
	\label{grasping_setup}
\end{figure}

In this section, effectiveness of the proposed contact force and torque estimation method for vision-based sensors is tested in grasping tasks. Sensing and visualization of contact information as well as adaptive control under external disturbances have been challenging tasks in robotic manipulation. Besides, situations are even more complex when introducing soft contact that brings in nonlinearity in deformation. Fig. \ref{friction_data} shows tangential force and normal force signals during multiple surface sliding trials (data collected by ATI nano-17 Force/Torque sensor). It is noticed from the chart on the upper right in Fig. \ref{friction_data} that the friction coefficient (equals to the ratio of tangential and normal forces $F_t/F_n$ when tangential force reaches each peak, as marked by blue circles) does not remain constant under different normal forces, which is one of the significant properties differences between hyperelastic contact and rigid contact. It also shows that within each trial of surface sliding, the ratio $F_t/F_n$ follows similar evolution: the ratio first rises; once reaches the maximum static friction coefficient, the ratio vibrates in a narrow band; the ratio drops afterwards, suggesting the occurrence of shear slip, as exhibited in the lower right chart of Fig. \ref{friction_data}.

\subsection{Grasping Stability Visualization}

\begin{figure}
	\centering
	\includegraphics[width=0.475\textwidth]{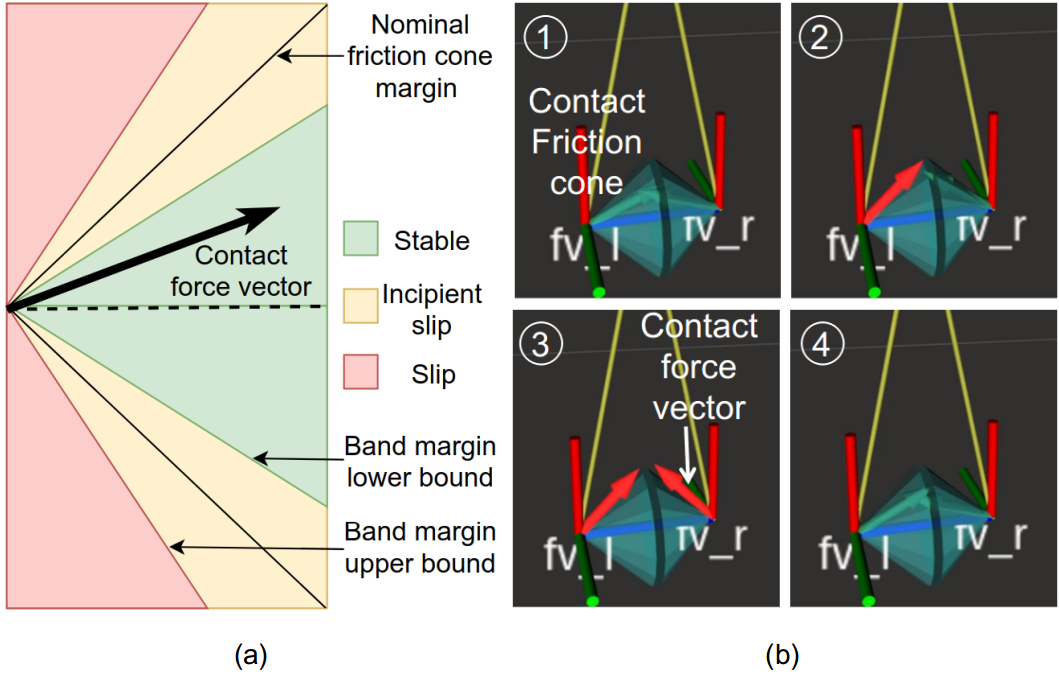}
	\caption{Schematic diagram of contact phases and visualization for experiment in Fig. \ref{grasping_setup}(a). Force signals are plotted in Fig. \ref{grasping_plots}(a-b).}
	\label{contact_phase_transitions}
\end{figure}

Taking behavior of $F_t/F_n$ during slip phases into consideration, we implement a visualization system for monitoring of grasping force and slip, as shown in Fig. \ref{contact_phase_transitions}. Since the friction coefficient is not constant, we take the average of friction coefficients across working range of normal force as the nominal value for simplification and visually illustrate friction cones \cite{mason2001mechanics} with this coefficient. The FingerVision sensors are installed on Robotiq 2-finger 140 gripper, serving as finger tips and sensing units, mimicking human fingertips. With the force and torque estimation module, we illustrate transitions of contact phases by classifying the spaces where contact force vectors reside w.r.t. the friction cones. As given in Fig. \ref{contact_phase_transitions}(a), contact statuses are classified into 4 phases: 1) Stable contact; 2) Incipient slip; 3) Slipping; 4) Recovery phase when force vector is regulated back into the yellow or green regions. In Fig. \ref{contact_phase_transitions}(b) contact forces are shown as arrows in green (when the vectors are within the friction cones) and red (outside of the  friction cones). The capability of indicating contact phases is beneficial to grasp reconfiguration for stable grasp.

\subsection{Feedback Control of Grasping Force}

In-hand manipulations of objects usually require minimal grasping forces, because the contact condition keeps switching between unstable and stable statuses, e.g. pen rolling in human hand. And when picking up fragile objects, power grasps also need to be avoided. Thus, adaptive control for grasping force is critical in many scenarios. Here we implement a simple feedback controller that takes in contact force estimation and maintains the ratios $F_t/F_n$ in a band on the peripheries of the friction cones (visualized with nominal friction coefficients as described previously). Details of the controller are given in Algorithm \ref{ag:controller}. In the algorithm, variables with subscripts \textit{l} and \textit{r} belong to the left and right contacts.  A conservative control strategy is implemented in our work. To maintain the contact forces in the vicinity of friction cone margins, the gripper decreases the opening if both left and right forces exceed the upper limits of band of the cones and increases the opening while both left and right forces are lower than the lower limits. 

\begin{algorithm}
	\caption{Grasping force controller}
	\hspace*{\algorithmicindent} \textbf{Input:} Contact forces  $F_l$, $F_r$; Gripper opening $D_g$;  \\
	\hspace*{\algorithmicindent} \quad \quad  \quad Band width $d$; Friction coefficient $\mu$.\\
	\hspace*{\algorithmicindent} \textbf{Output:} Gripper requested opening $D_r$.
	\label{pseudoPSO}
	\begin{algorithmic}[1]
		\State Initialize $r_l$, $r_r$ with $\mu$
		\While {True}
		\State $r_l \gets F_{lt}/F_{ln}$, $r_r \gets F_{rt}/F_{rn}$
		
		\If{$r_l > \mu+d/2$ \textbf{and} $r_r > \mu+d/2$ }
		\State $D_r = D_g - 1$
		\EndIf
		
		\If{$r_l < \mu+d/2$ \textbf{and} $r_r < \mu+d/2$ }
		\State $D_r = D_g + 1$
		\EndIf
		
		\EndWhile
	\end{algorithmic}
	\label{ag:controller}
\end{algorithm}

\begin{figure}
	\centering
	\includegraphics[width=0.485\textwidth]{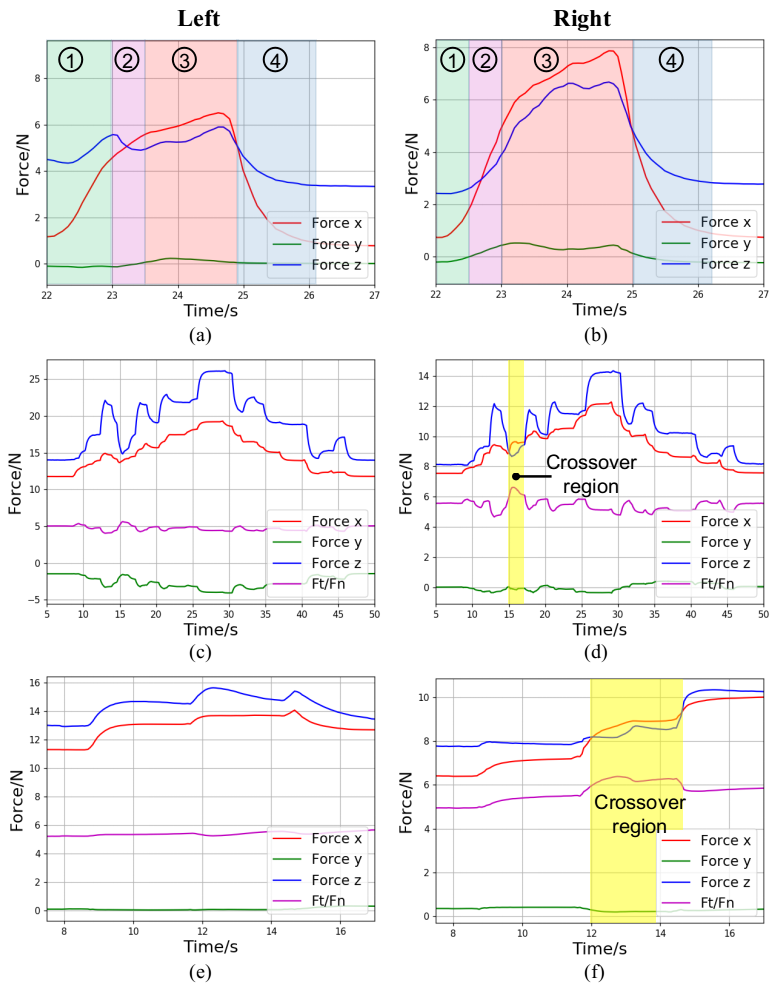}
	\caption{Grasping contact force signals under loads. Force x, y and z are the projections of the contact force onto the sensor surface coordinate system in Fig. \ref{calibration_setup}. (a-b). Changes of contact forces for manual press on grasped object, with constant opening distance between two fingertips. Four contact phases are illustrated in different colors. (c-d). Contact force signals during loading and unloading, with active grasp force controller. (e-f). Contact force signals during loading, without  grasping force controller. The unloading process is not given since contacts are broken, which leads to grasping failure.}
	\label{grasping_plots}
\end{figure}

The controller performs well in maintaining stable contact using minimal grasping forces in the object holding experiment during loading and unloading of weights that result in increase and decrease of tangential forces. As demonstrated in Fig. \ref{grasping_plots}(c-d), with controller being active, there are no or much shorter periods of crossovers(indicated as periods when $F_t/F_n$ dramatically rises that leads to contact slip). The ratio $F_t/F_n$ in right-side fingertip recovers quickly from the crossover region (rendered in yellow in Fig. \ref{grasping_plots}) due to the regulation of the force controller. The regulation process can also be seen from the visualization system, which keeps the contact force vectors around the margins of friction cones. Without force control, there is an extended longer period of crossover during loading process. The gripper fails to maintain contact forces inside the friction cones. Grasp fails if at least one contact breaks. It is worth noting that after slip happens on the surface, another crossover occurs due to the fact that dynamic friction coefficient is lower than static friction coefficient. The signals of left and right sensors are not of exactly the same forms, which could stem from the variances in sensor fabrication and calibration, object alignment difference for two contact surfaces and gripper pose not being exactly upright that leads to imbalanced loads on two fingertips.

\section{Conclusion}

In this work, we develop a contact force and torque estimation method for vision-based tactile sensor using Helmholtz-Hodge Decomposition (HHD). Starting from observations of the relations between contact force and torque and marker displacement patterns, we establish the mapping from decomposed components of HHD to contact force and torque estimation. In characteristic experiment, the force and torque estimation results show high linearity and guarantee lower demands for data size and better accuracy on predictions using models with low complexity. The proposed method is further tested in both contact stability visualization and grasping with adaptive force control for verification of effectiveness and presents potential in facilitating studies of grasping stability metric. Future works fall mainly on integrating sensor and algorithm into grasping system to predict high-level physical information including object center of mass, estimation of object dynamics, and prediction of grasping stability.


\bibliographystyle{IEEEtran}
\bibliography{reference}

\end{document}